\documentclass[10pt,twocolumn,letterpaper]{article}

\usepackage{cvpr}
\usepackage{times}
\usepackage{epsfig}
\usepackage{graphicx}
\usepackage{amsmath}
\usepackage{amssymb}

\usepackage{subeqnarray}
\usepackage{color}
\usepackage{amstext}

\usepackage{array}

\usepackage{subfigure}
\usepackage{setspace}
\usepackage[noblocks]{authblk}
\usepackage{chngpage}
\usepackage{multirow}
\graphicspath{{figure/},{figure1/},{figure2/},{figure3/}}

\usepackage[breaklinks=true,bookmarks=false]{hyperref}

\cvprfinalcopy 

\def\cvprPaperID{****} 

\begin{document}

\title{Multi-level Wavelet-CNN for Image Restoration}
\author[1]{Pengju Liu}
\author[1]{Hongzhi Zhang \thanks{Corresponding author.}}
\author[1]{Kai Zhang}
\author[2]{Liang Lin}
\author[1]{Wangmeng Zuo}
\affil[1]{\normalsize School of Computer Science and Technology, Harbin Institute of Technology, China}
\affil[2]{\normalsize School of Data and Computer Science, Sun Yat-Sen University, Guangzhou, China
\authorcr{\tt \footnotesize{lpj008@126.com, zhanghz0451@gmail.com, linliang@ieee.org, cskaizhang@gmail.com, cswmzuo@gmail.com}}}




\maketitle
\pagestyle{empty}
\thispagestyle{empty}

\begin{abstract}
The tradeoff between receptive field size and efficiency is a crucial issue in low level vision.
Plain convolutional networks (CNNs) generally enlarge the receptive field at the expense of computational cost.
%
%
Recently, dilated filtering has been adopted to address this issue.
But it suffers from gridding effect, and the resulting receptive field is only a sparse sampling of input image with checkerboard patterns.
%
%
In this paper, we present a novel multi-level wavelet CNN (MWCNN) model for better tradeoff between receptive field size and computational efficiency.
With the modified U-Net architecture, wavelet transform is introduced to reduce the size of feature maps in the contracting subnetwork.
Furthermore, another convolutional layer is further used to decrease the channels of feature maps.
In the expanding subnetwork, inverse wavelet transform is then deployed to reconstruct the high resolution feature maps.
Our MWCNN can also be explained as the generalization of dilated filtering and subsampling, and can be applied to many image restoration tasks.
The experimental results clearly show the effectiveness of MWCNN for image denoising, single image super-resolution, and JPEG image artifacts removal.
\end{abstract}

\section{Introduction}
Image restoration, which aims to recover the latent clean image $\mathbf{x}$ from its degraded observation $\mathbf{y}$, is a fundamental and long-standing problem in low level vision.
For decades, varieties of methods have been proposed for image restoration from both prior modeling and discriminative learning perspectives~\cite{banham1997digital, katsaggelos2012digital, Chen2015Trainable, dabov2007image, gu2014weighted, schmidt2014shrinkage, wright2009robust}.
Recently, convolutional neural networks (CNNs) have also been extensively studied and {\color{black} achieved} state-of-the-art performance in several representative image restoration tasks, such as single image super-resolution (SISR)~\cite{dong2016image,kim2015accurate,Ledig2017Photo}, image denoising~\cite{Zhang2016Beyond}, image deblurring~\cite{zhang2017learning}, and lossy image compression~\cite{Li2017Learning}.
The popularity of CNN in image restoration can be explained from two aspects.
On the one hand, existing CNN-based solutions have outperformed the other methods with a large margin for several simple tasks such as image denoising and SISR~\cite{dong2016image,kim2015accurate,Ledig2017Photo,Zhang2016Beyond}.
On the other hand, recent studies have revealed that one can plug CNN-based denoisers into model-based optimization methods for solving more complex image restoration tasks~\cite{romano2016little, zhang2017learning}, which also promotes the widespread use of CNNs.

\begin{figure}[!t]
\vspace{-2ex}
\begin{center}
  \vspace{-0.5ex}
  \includegraphics[width=0.41\textwidth]{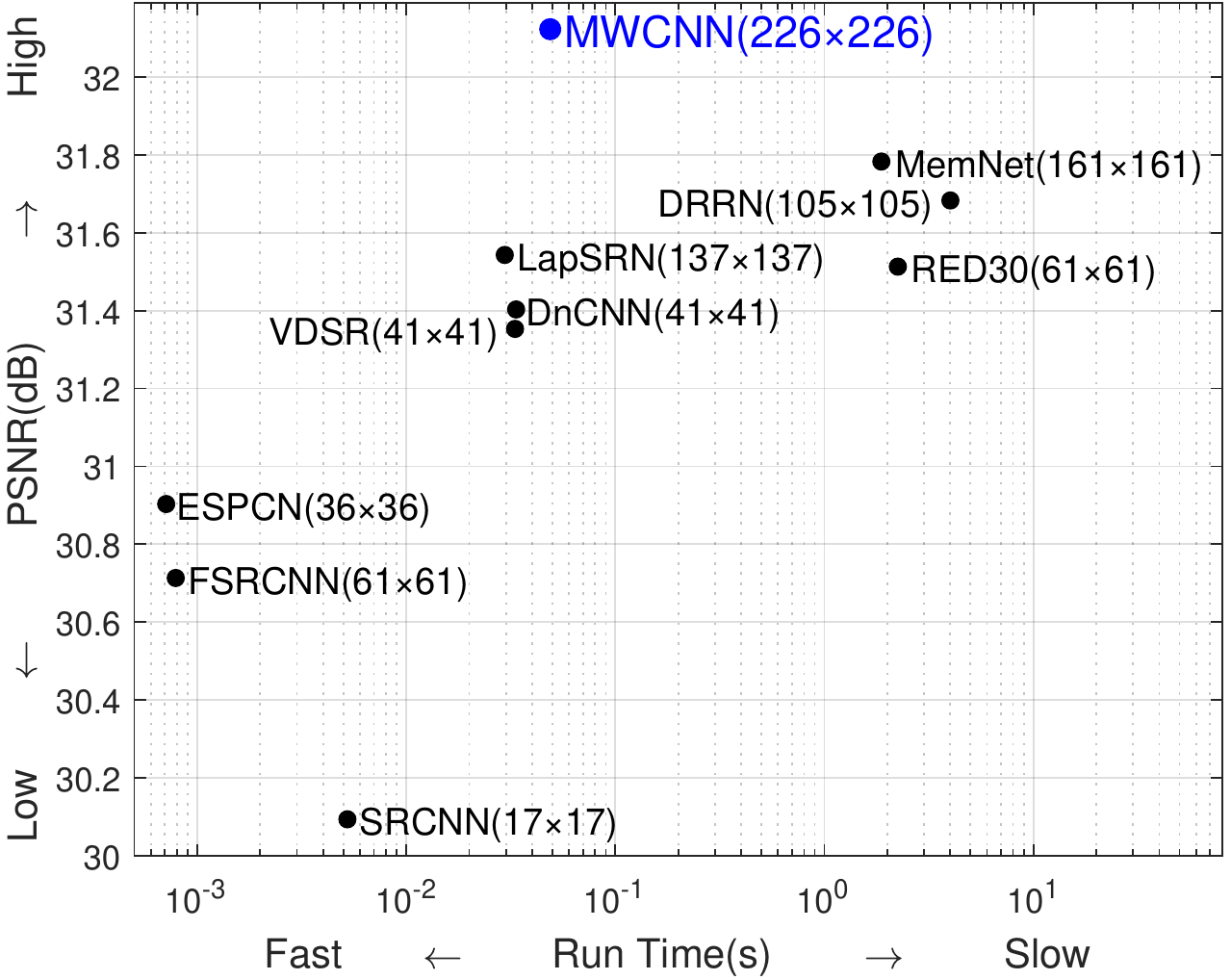}\\
  \vspace{-2.6ex}
\end{center}
  \caption{The run time vs. PSNR value of representative CNN models, including SRCNN~\cite{dong2016image}, FSRCNN~\cite{Dong2016Accelerating}, ESPCN~\cite{Shi2016Real}, VDSR~\cite{kim2015accurate}, DnCNN~\cite{Zhang2016Beyond}, RED30~\cite{Mao2016Image}, LapSRN~\cite{lai2017deep}, DRRN~\cite{Tai2017Image}, MemNet~\cite{Tai2017Image} and our MWCNN. The receptive field of each model are also provided. The PSNR and time are evaluated on Set5 with the scale factor $\times4$ running on a GTX1080 GPU.}
  \vspace{-3ex}
  \label{fig:receptive}
\end{figure}

For image restoration, CNN actually represents a mapping from degraded observation to latent clean image.
Due to the input and output images usually should be of the same size, one representative strategy is to use the fully convolutional network (FCN) by removing the pooling layers.
In general, larger receptive field is helpful to restoration performance by taking more spatial context into account.
However, for FCN without pooling, the receptive field size can be enlarged by either increasing the network depth or using {\color{black} filters with larger size}, which unexceptionally results in higher computational cost.
%
%
In~\cite{zhang2017learning}, dilated filtering~\cite{yu2015multi} is adopted to enlarge receptive field without the sacrifice of computational cost.
Dilated filtering, however, inherently suffers from gridding effect~\cite{Wang2017Understanding}, where the receptive field only considers a sparse sampling of input image with checkerboard patterns.
%
%
%
%
%
Thus, one should be careful to enlarge receptive field while avoiding the increase of computational burden and the potential sacrifice of performance improvement.
Taking SISR as an example, Figure~\ref{fig:receptive} illustrates the receptive field, run times, and PSNR values of several representative CNN models.
It can be seen that FSRCNN~\cite{Dong2016Accelerating} has relatively larger receptive field but achieves lower PSNR value than VDSR~\cite{kim2015accurate} and DnCNN~\cite{Zhang2016Beyond}. 

In this paper, we present a multi-level wavelet CNN (MWCNN) model to enlarge receptive field for better tradeoff between performance and efficiency.
Our MWCNN is based on the U-Net~\cite{Ronneberger2015U} architecture consisting of a contracting subnetwork and an expanding subnetwork.
In the contracting subnetwork, discrete wavelet transform (DWT) is introduced to replace each pooling operation.
Since DWT is invertible, it is guaranteed that all the information can be kept by such downsampling scheme.
Moreover, DWT can capture both frequency and location information of feature maps~\cite{daubechies1990wavelet,daubechies1992ten}, which may be helpful in preserving detailed texture.
%
%
In the expanding subnetwork, inverse wavelet transform (IWT) is utilized for upsampling low resolution feature maps to high resolution ones.
To enrich feature representation and reduce computational burden, element-wise summation is adopted for combining the feature maps from the contracting and expanding subnetworks.
Moreover, dilated filtering can also be explained as a special case of MWCNN, and ours is more general and effective in enlarging receptive field.
Experiments on image denoising, SISR, and JPEG image artifacts removal validate the effectiveness and efficiency of our MWCNN.
{\color{black} As shown in Figure~\ref{fig:receptive}, MWCNN is moderately slower than LapSRN~\cite{lai2017deep}, DnCNN~\cite{Zhang2016Beyond} and VDSR~\cite{kim2015accurate} in terms of run time, but can have a much larger receptive field and higher PSNR value.}
To sum up, the contributions of this work include:
\vspace{-0.025in}
\begin {itemize}
   \item A novel MWCNN model to enlarge receptive field with better tradeoff between efficiency and restoration performance.  \vspace{-0.025in}
   \item Promising detail preserving ability due to the good time-frequency localization of DWT. \vspace{-0.025in}
   \item State-of-the-art performance on image denoising, SISR, and JPEG image artifacts removal. \vspace{-0.025in}
 \end {itemize}

\section{Related work}\label{sec:related}

In this section, we present a brief review on the development of CNNs for image denoising, SISR, JPEG image artifacts removal, and other image restoration tasks.
Specifically, more discussions are given to the relevant works on enlarging receptive field and incorporating DWT in CNNs.

\subsection{Image denoising}

Since 2009, CNNs have been applied for image denoising~\cite{jain2009natural}.
These early methods generally cannot achieve state-of-the-art denoising performance~\cite{Agostinelli2013Robust,jain2009natural,Xie2012Image}.
Recently, multi-layer perception (MLP) has been adopted to learn the mapping from noise patch to clean pixel, and achieve comparable performance with BM3D~\cite{burger2012image}.
By incorporating residual learning with batch normalization~\cite{ioffe2015batch}, the DnCNN model by Zhang \etal~\cite{Zhang2016Beyond} can outperform traditional non-CNN based methods.
Mao \etal~\cite{Mao2016Image} suggest to add symmetric skip connections to FCN for improving denoising performance.
For better tradeoff between speed and performance, Zhang \etal~\cite{zhang2017learning} present a 7-layer FCN with dilated filtering.
Santhanam \etal~\cite{santhanam2017generalized} introduce a recursively branched deconvolutional network (RBDN), where pooling/unpooling is adopted to obtain and aggregate multi-context representation.

\subsection{Single image super-resolution}

The application of CNN in SISR begins with SRCNN~\cite{dong2016image}, which adopts a 3-layer FCN without pooling and has a small receptive field.
Subsequently, very deep network~\cite{kim2015accurate}, residual units~\cite{Ledig2017Photo}, Laplacian pyramid~\cite{lai2017deep},
and recursive architecture~\cite{kim2016deeply,Tai2017Image} have also been suggested to enlarge receptive field.
These methods, however, enlarge the receptive field {\color{black} at the cost of either} increasing computational cost or loss of information.
Due to the speciality of SISR, one effective approach is to take the low-resolution (LR) image as input to CNN~\cite{Dong2016Accelerating,Shi2016Real} for better tradeoff between receptive field size and efficiency.
In addition, generative adversarial networks (GANs) have also been introduced to improve the visual quality of SISR~\cite{johnson2016perceptual,Ledig2017Photo,sajjadi2016enhancenet}.

\subsection{JPEG image artifacts removal}
Due to high compression rate, JPEG image usually suffers from blocking effect and results in unpleasant visual quality.
In~\cite{Dong2016Compression}, Dong \etal adopt a 4-layer ARCNN for JPEG image deblocking.
By taking the degradation model of JPEG compression into account~\cite{Chen2015Trainable,Wang2016D3}, Guo \etal~\cite{Guo2016Building} suggest a dual-domain convolutional network to combine the priors in both DCT and pixel domains.
GAN has also been introduced to generate more realistic result~\cite{Guo2017One}. 

\subsection{Other restoration tasks}

Due to the similarity of image denoising, SISR, and JPEG artifacts removal, the model suggested for one task may be easily extended to the other tasks simply by retraining.
For example, both DnCNN~\cite{Zhang2016Beyond} and MemNet~\cite{tai2017memnet} have been evaluated on all the three tasks.
Moreover, CNN denoisers can also serve as a kind of plug-and-play prior.
By incorporating with unrolled inference, any restoration tasks can be tackled by sequentially applying the CNN denoisers~\cite{zhang2017learning}.
Romano \etal~\cite{romano2016little} further propose a regularization by denoising framework, and provide an explicit functional for defining the regularization induced by denoisers.
These methods not only promote the application of CNN in low level vision, but also present many solutions to exploit CNN denoisers for other image restoration tasks.

Several studies have also been given to incorporate wavelet transform with CNN.
Bae \etal~\cite{bae2017beyond} find that learning CNN on wavelet subbands benefits CNN learning, and suggest a wavelet residual network (WavResNet) for image denoising and SISR.
Similarly, Guo \etal~\cite{guo2017deep} propose a deep wavelet super-resolution (DWSR) method to recover missing details on subbands.
Subsequently, deep convolutional framelets \cite{Han2017Framing, Ye2017Deep} {\color{black}have been} developed to extend convolutional framelets for low-dose CT.
However, both of WavResNet and DWSR only consider one level wavelet decomposition.
Deep convolutional framelets independently processes each subband from decomposition perspective, which ignores the dependency between these subbands.
%
In contrast, multi-level wavelet transform is considered by our MWCNN to enlarge receptive field without information loss.
Taking all the subbands as inputs after each transform, our MWCNN can embed DWT to any CNNs with pooling, and owns more power to model both spatial context and inter-subband dependency.
%

\section{Method}\label{sec:method}

In this section, we first introduce the multi-level wavelet packet transform (WPT).
Then we present our MWCNN motivated by multi-level WPT, and describe its network architecture. 
Finally, discussion is given to analyze the connection of MWCNN with dilated filtering and subsampling.

\subsection{From multi-level WPT to MWCNN}


In 2D discrete wavelet transform (DWT), four filters, i.e. $\mathbf{f}_{LL}$, $\mathbf{f}_{LH}$, $\mathbf{f}_{HL}$, and $\mathbf{f}_{HH}$, are used to convolve with an image $\mathbf{x}$~\cite{mallat1989theory}.
The convolution results are then downsampled to obtain the four subband images $\mathbf{x}_{1}$, $\mathbf{x}_{2}$, $\mathbf{x}_{3}$, and $\mathbf{x}_{4}$.
%
%
For example, $\mathbf{x}_{1}$ is defined as $(\mathbf{f}_{LL} \otimes \mathbf{x})\downarrow_2$.
Even though the downsampling operation is deployed, due to the biorthogonal property of DWT, the original image $\mathbf{x}$ can be accurately reconstructed by the inverse wavelet transform (IWT), \ie, $\mathbf{x} = IWT(\mathbf{x}_{1}, \mathbf{x}_{2}, \mathbf{x}_{3}, \mathbf{x}_{4})$.

\begin{figure}[!t]
\begin{center}
\vspace{-1.0ex}
\hspace{-0ex}
\subfigure[Multi-level WPT architecture]{
\begin{minipage}[c]{0.49\textwidth}
\centering
  \includegraphics[width=0.98\linewidth]{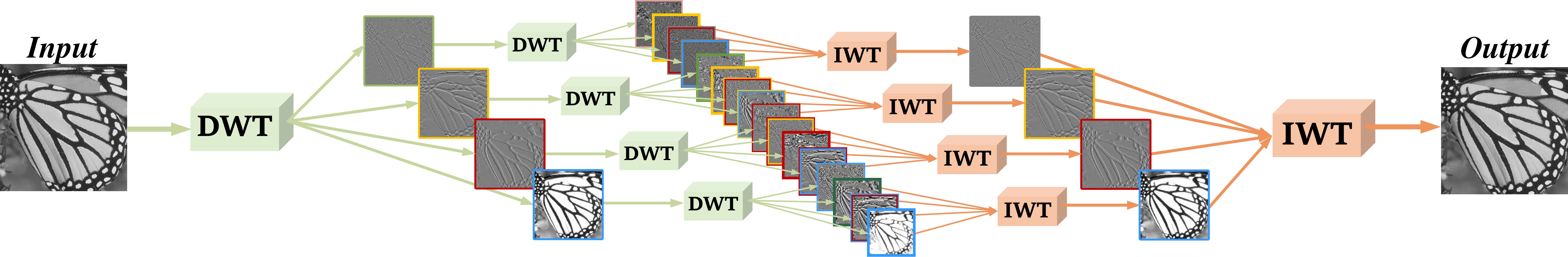}
  \label{fig:wpt_arch}
\end{minipage}%
}
\vspace{-0.8ex}

\vspace{-1.0ex}
\subfigure[Multi-level wavelet-CNN architecture]{
\begin{minipage}[c]{0.49\textwidth}
\centering
  \includegraphics[width=0.98\linewidth]{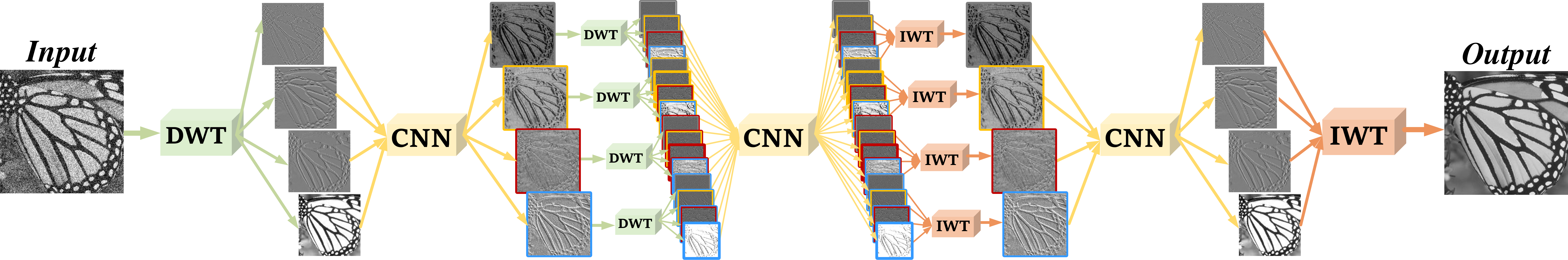}
  \label{fig:mwcnn_arch}
\end{minipage}%
}
\vspace{-3.0ex}
\end{center}
  \caption{From WPT to MWCNN.
  Intuitively, WPT can be seen as a special case of our MWCNN without CNN blocks.
  }
 \vspace{-2ex}

 \label{fig:wpt}
\end{figure}

In multi-level wavelet packet transform (WPT)~\cite{akansu2001multiresolution,daubechies1992ten}, the subband images $\mathbf{x}_{1}$, $\mathbf{x}_{2}$, $\mathbf{x}_{3}$, and $\mathbf{x}_{4}$ are further processed with DWT to produce the decomposition results.
For two-level WPT, each subband image $\mathbf{x}_{i}$ ($i =$ 1, 2, 3, or $4$) is decomposed into four subband images $\mathbf{x}_{i,1}$, $\mathbf{x}_{i,2}$, $\mathbf{x}_{i,3}$, and $\mathbf{x}_{i,4}$.
Recursively, the results of three or higher levels WPT can be attained.
Figure \ref{fig:wpt_arch} illustrates the decomposition and reconstruction of an image with WPT.
Actually, WPT is a special case of FCN without the nonlinearity layers.
In the decomposition stage, four pre-defined filters are deployed to each (subband) image, and downsampling is then adopted as the pooling operator.
In the reconstruction stage, the four subband images are first upsampled and then convolved with the corresponding filters to produce the reconstruction result at the current level.
Finally, the original image $\mathbf{x}$ can be accurately reconstructed by inverse WPT.

In image denoising and compression, some operations, \eg, soft-thresholding and quantization, usually are required to process the decomposition result~\cite{chang2000adaptive,lewis1992image}.
These operations can be treated as some kind of nonlinearity tailored to specific task.
In this work, we further extend WPT to multi-level wavelet-CNN (MWCNN) by adding a CNN block between any two levels of DWTs, as illustrated in Figure \ref{fig:mwcnn_arch}.
After each level of transform, all the subband images are taken as the inputs to a CNN block to learn a compact representation as the inputs to the subsequent level of transform.
It is obvious that MWCNN is a generalization of multi-level WPT, and degrades to WPT when each CNN block becomes the identity mapping.
Due to the biorthogonal property of WPT, our MWCNN can use subsampling operations safely without information loss.
Moreover, compared with conventional CNN, the frequency and location characteristics of DWT is also expected to benefit the preservation of detailed texture.

\begin{figure*}[!t]
\begin{center}
\vspace{-2.6ex}
  \includegraphics[width=0.9\textwidth]{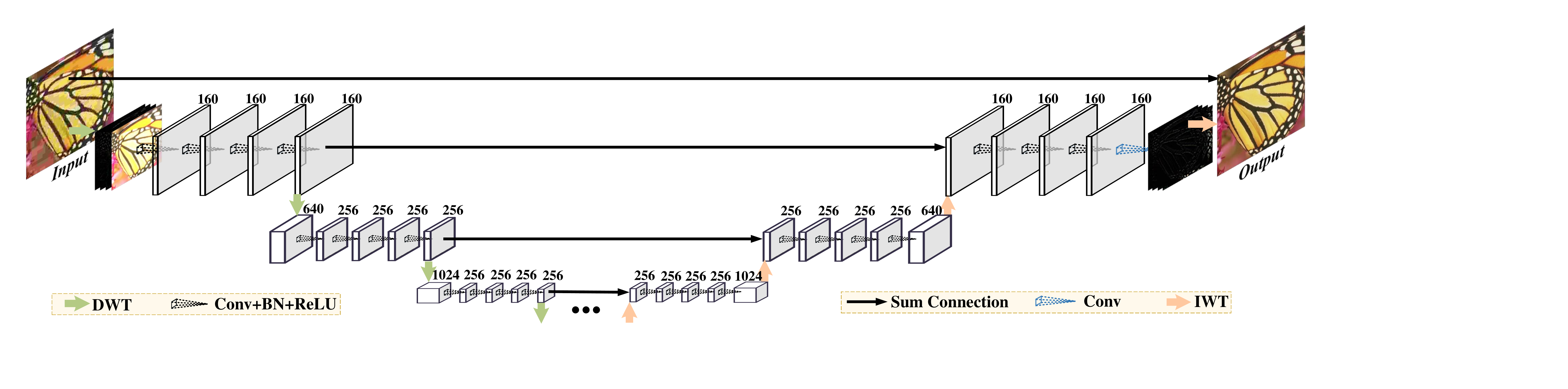}\\
  \vspace{-2ex}
\end{center}
  \caption{Multi-level wavelet-CNN architecture.
  It consists two parts: the contracting and expanding subnetworks.
  Each solid box corresponds to a multi-channel feature map.
  And the number of channels is annotated on the top of the box.
  The network depth is 24. 
  Moreover, our MWCNN can be further extended to higher level (e.g., $\geq 4$) by duplicating the configuration of the 3rd level subnetwork.
  }
  \vspace{-2ex}
  \label{fig:Architecture}
\end{figure*}

\subsection{Network architecture}
\label{sec3.2}

The key of our MWCNN architecture is to design the CNN block after each level of DWT.
As shown in Figure~\ref{fig:Architecture}, each CNN block is a 4-layer FCN without pooling, and takes all the subband images as inputs.
In contrast, different CNNs are deployed to low-frequency and high-frequency bands in deep convolutional framelets~\cite{Han2017Framing,Ye2017Deep}.
We note that the subband images after DWT are still dependent, and the ignorance of their dependence may be harmful to the restoration performance.
Each layer of the CNN block is composed of convolution with $3 \times 3$ filters (Conv), batch normalization (BN), and rectified linear unit (ReLU) operations.
As to the last layer of the last CNN block, Conv without BN and ReLU is adopted to predict residual image.

Figure~\ref{fig:Architecture} shows the overall architecture of MWCNN which consists of a contracting subnetwork and an expanding subnetwork.
Generally, MWCNN modifies U-Net from three aspects.
(i) For downsampling and upsampling, max-pooling and up-convolution are used in conventional U-Net\cite{Ronneberger2015U}, while DWT and IWT are utilized in MWCNN.
(ii) For MWCNN, the downsampling results in the increase of feature map channels.
Except the first one, the other CNN blocks are deployed to reduce the feature map channels for compact representation.
In contrast, for conventional U-Net, the downsampling has no effect on feature map channels, and the subsequent convolution layers are used to increase feature map channels.
(iii) In MWCNN, element-wise summation is used to combine the feature maps from the contracting and expanding subnetworks.
While in conventional U-Net concatenation is adopted.
{\color{black} Then our final network contains 24 layers.}
For more details on the setting of MWCNN, please refer to Figure~\ref{fig:Architecture}.
In our implementation, Haar wavelet is adopted as the default in MWCNN.
Other wavelets, \eg, Daubechies 2 (DB2), are also considered in our experiments.

Denote by $\Theta$ the network parameters of MWCNN, and $F\left(\mathbf{y}; \Theta \right)$ be the network output.
Let $\{\left( \mathbf{y}_i, \mathbf{x}_i \right)\}_{i=1}^N$ be a training set, where $ \mathbf{y}_i$ is the $i$-th input image, $ \mathbf{x}_i $ is the corresponding ground-truth image.
The objective function for learning MWCNN is then given by
\begin{equation}\label{eq:loss}
  \mathcal{L}(\Theta) = \frac{1}{2N}\sum_{i=1}^N \|F(\mathbf{y}_i; \Theta)  - \mathbf{x}_i\|_F^2.
\end{equation}
The ADAM algorithm~\cite{kingma2014adam} is adopted to train MWCNN by minimizing the objective function.
Different from VDSR~\cite{kim2015accurate} and DnCNN~\cite{Zhang2016Beyond}, we do not adopt the residual learning formulation for the reason that it can be naturally embedded in MWCNN.

\subsection{Discussion}

The DWT in MWCNN is closely related with the pooling operation and dilated filtering.
By using the Haar wavelet as an example, we explain the connection between DWT and sum-pooling.
In 2D Haar wavelet, the low-pass filter $\mathbf{f}_{LL}$ is defined as,
\begin{equation}\label{eq:haar_ll}
  \mathbf{f}_{LL} = \begin{bmatrix}
                    1 & 1\\
                    1 & 1
                    \end{bmatrix}.
\end{equation}
One can see that $(\mathbf{f}_{LL} \otimes \mathbf{x})\downarrow_2$ actually is the sum-pooling operation.
When only the low-frequency subband is considered, DWT and IWT will play the roles of pooling and up-convolution in MWCNN, respectively.
When all the subbands are taken into account, MWCNN can avoid the information loss caused by conventional subsampling, and may benefit restoration result.

To illustrate the connection between MWCNN and dilated filtering with factor 2, we first give the definition of $\mathbf{f}_{LH}$, $\mathbf{f}_{HL}$, and $\mathbf{f}_{HH}$,
\begin{equation}\label{eq:haar_other}
  \mathbf{f}_{LH} \!=\! \begin{bmatrix}
                    -1 \!\!&\!\! -1\\
                    1 \!\!&\!\! 1
                    \end{bmatrix},
    \mathbf{f}_{HL} \!=\! \begin{bmatrix}
                    -1 \!\!&\!\! 1\\
                    -1 \!\!&\!\! 1
                    \end{bmatrix},
      \mathbf{f}_{HH} \!=\! \begin{bmatrix}
                    1 \!\!&\!\! -1\\
                    -1 \!\!&\!\! 1
                    \end{bmatrix}.
\end{equation}
Given an image $\mathbf{x}$ with size of $m \times n$, the $(i,j)$-th value of $\mathbf{x}_1$ after 2D Haar transform can be written as $\mathbf{x}_1(i,j) = \mathbf{x}(2i-1,2j-1) + \mathbf{x}(2i-1,2j) + \mathbf{x}(2i,2j-1) + \mathbf{x}(2i,2j)$.
And $\mathbf{x}_2(i,j)$, $\mathbf{x}_3(i,j)$, and $\mathbf{x}_4(i,j)$ can be defined analogously.
We also have {\color{black}$\mathbf{x}(2i-1,2j-1) = \left(\mathbf{x}_1(i,j) - \mathbf{x}_2(i,j) - \mathbf{x}_3(i,j) + \mathbf{x}_4(i,j)\right)/4$}.
The dilated filtering with factor 2 on the position $(2i-1,2j-1)$ of $\mathbf{x}$ can be written as
\begin{equation}\label{eq:dilated}
  (\mathbf{x} \otimes_2 \mathbf{k})(2i-1,2j-1) =  \sum_{ \!\!\! \! \! \!\!
  \begin{scriptsize}
  \begin{aligned}
    p\!+\!2s\! = \!2i\!-\!1,\\
    q\!+\!2t\! = \!2j\!-\!1
  \end{aligned}
  \end{scriptsize}
} \mathbf{x}(p,q) \mathbf{k}(s,t),
\end{equation}
where $\mathbf{k}$ is the $3 \times 3$ convolution kernel.
Actually, it also can be obtained by using the $3 \times 3$ convolution with the subband images,
\begin{equation}\label{eq:dilated_haar}
  (\mathbf{x} \otimes_2 \mathbf{k})(2i\! -\! 1,2j\! -\! 1) \! \! = \! \! \left((\mathbf{x}_1 \!\! - \!\! \mathbf{x}_2 \!\! - \!\! \mathbf{x}_3 \!\! + \!\! \mathbf{x}_4) \! \otimes \! \mathbf{k}\right)(i,j)/4.
\end{equation}
Analogously, we can analyze the connection between dilated filtering and MWCNN for $(\mathbf{x} \otimes_2 \mathbf{k})(2i-1,2j)$, $(\mathbf{x} \otimes_2 \mathbf{k})(2i,2j-1)$, $(\mathbf{x} \otimes_2 \mathbf{k})(2i,2j)$.
Therefore, the $3 \times 3$ dilated convolution on $\mathbf{x}$ can be treated as a special case of $4 \times 3 \times 3$ convolution on the subband images.

\begin{figure}[!htbp]
\scriptsize{
\vspace{-2ex}
\begin{center}
\subfigure[]{
\begin{minipage}[c]{0.133\textwidth}
\centering
  \includegraphics[width=0.99\linewidth]{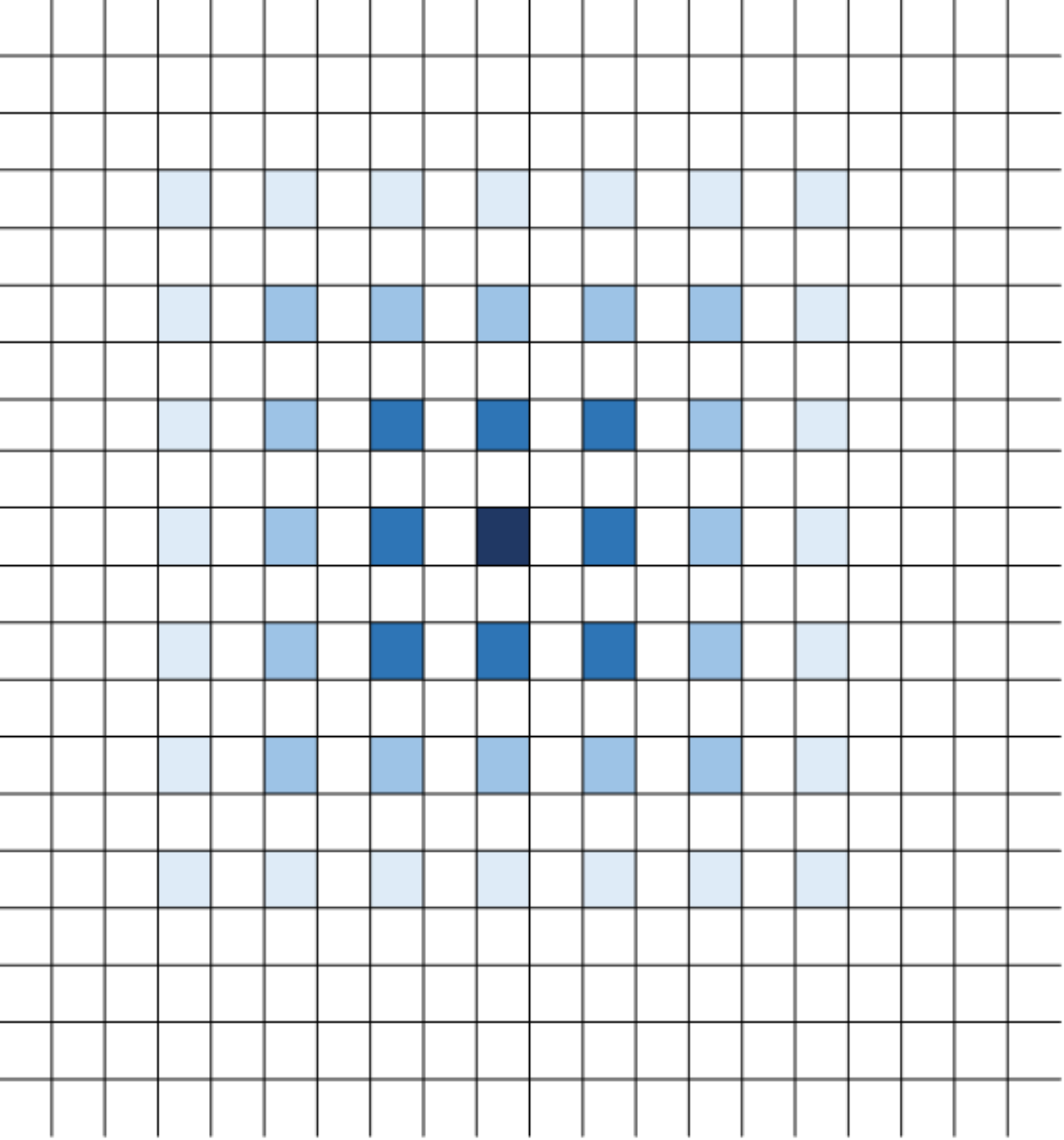}
  \label{fig:RF_d}
\end{minipage}%
}
\hspace{2.6ex}
\subfigure[]{
\begin{minipage}[c]{0.133\textwidth}
\centering
  \includegraphics[width=0.99\linewidth]{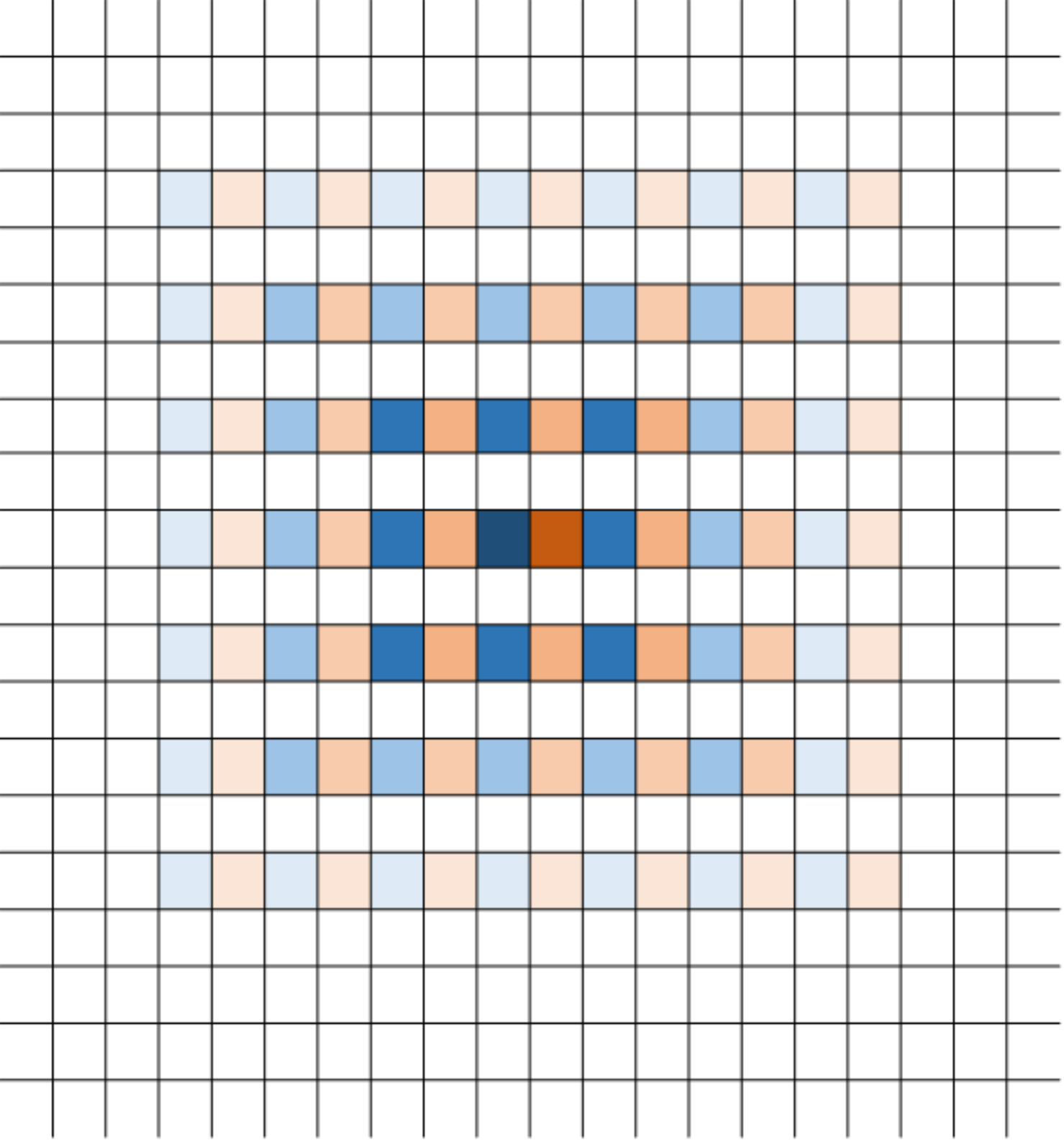}
  \label{fig:RF2D}
\end{minipage}%
}
\hspace{2.6ex}
\subfigure[]{
\begin{minipage}[c]{0.133\textwidth}
\centering
  \includegraphics[width=0.99\linewidth]{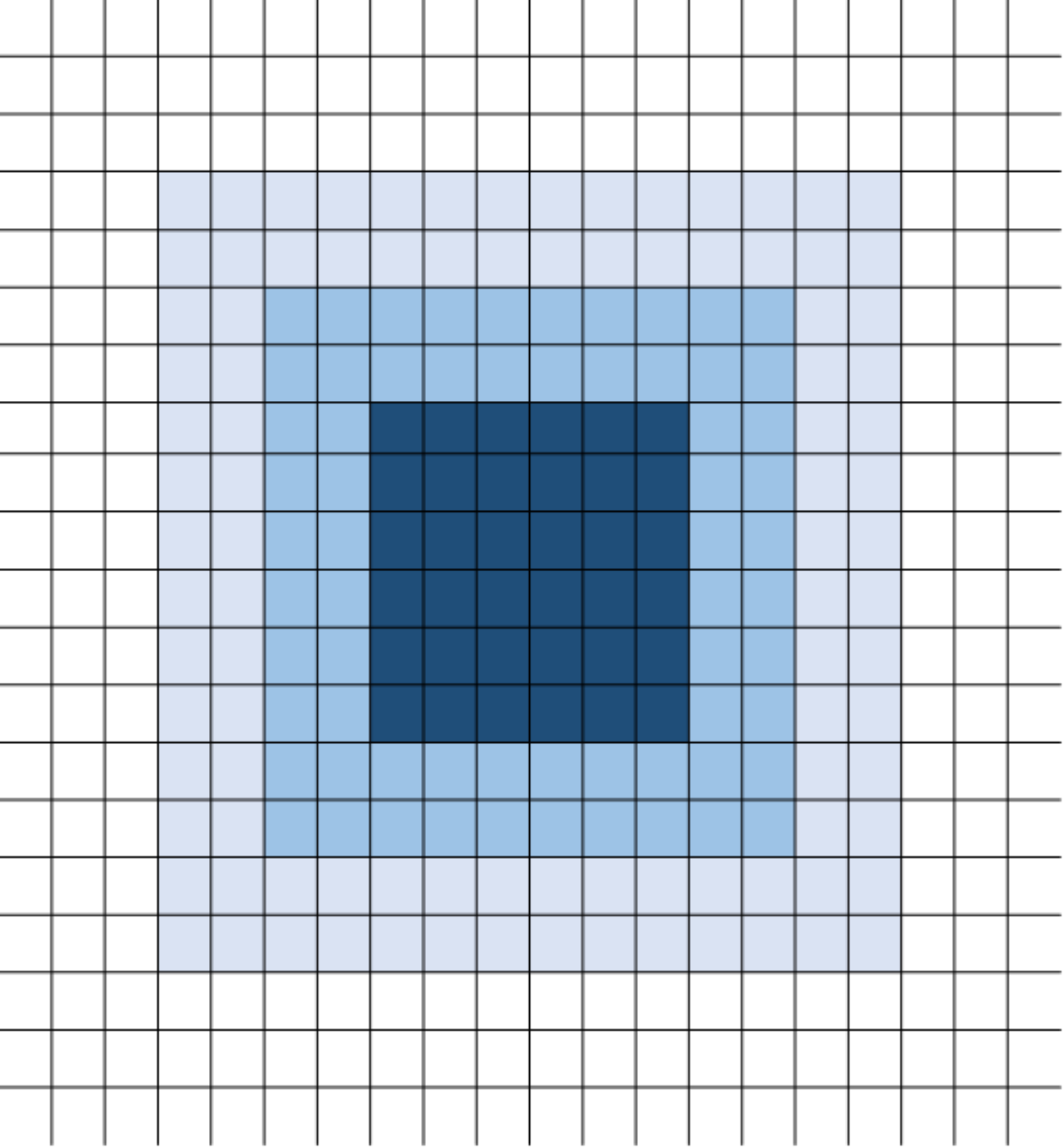}
  \label{fig:RFMW}
\end{minipage}%
}
\caption{\small Illustration of the gridding effect.
         Taken 3-layer CNNs as an example: (a) the dilated filtering with factor 2 surfers large portion of information loss, (b) and the two neighbored pixels are based on information from totally non-overlapped locations, (c) while our MWCNN can perfectly avoid underlying drawbacks.
}\label{fig:gridding}
\vspace{-5ex}
\end{center}}
\end{figure}

Compared with dilated filtering, MWCNN can also avoid the gridding effect.
%
%
After several layers of dilated filtering, it only considers a sparse sampling of locations with the checkerboard pattern, resulting in large portion of information loss (see Figure~\ref{fig:RF_d}).
Another problem with dilated filtering is that the two neighbored pixels may be based on information from totally non-overlapped locations (see Figure~\ref{fig:RF2D}), and may cause the inconsistence of local information.
In contrast, Figure~\ref{fig:RFMW} illustrates the receptive field of MWCNN.
One can see that MWCNN is able to well address the sparse sampling and inconsistence of local information, and is expected to benefit restoration performance quantitatively and qualitatively.


\section{Experiments}
\label{sec:exp}

{\color{black}
%
Experiments are conducted for performance evaluation on three tasks, \ie, image denoising, SISR, and compression artifacts removal.
Comparison of several MWCNN variants is also given to analyze the contribution of each component.
}
{\color{black}The code and pre-trained models will be given at \url{https://github.com/lpj0/MWCNN}}.

\subsection{Experimental setting}

\subsubsection{Training set}
To train our MWCNN, a large training set is constructed by using images from three dataset, \ie Berkeley Segmentation Dataset (BSD)~\cite{MartinFTM01}, DIV2K~\cite{agustsson2017ntire} and Waterloo Exploration Database (WED)~\cite{Ma2017Waterloo}.
Concretely, we collect $200$ images from BSD, $800$ images from DIV2K, and $4,744$ images from WED.
Due to the receptive field of MWCNN is not less than $226 \times 226$, in the training stage $N = 24 \times 6,000$  patches with the size of $240\times 240$ are cropped from the training images.

For image denoising, Gaussian noise with specific noise level is added to clean patch, and MWCNN is trained to learn a mapping from noisy image to denoising result.
Following \cite{Zhang2016Beyond}, we consider three noise levels, \ie, $\sigma$ = 15, 25 and 50.
For SISR, we take the result by bicubic upsampling as the input to MWCNN, and three specific scale factors, \ie., $\times2$, $\times3$ and $\times4$, are considered in our experiments.
For JPEG image artifacts removal, we follow \cite{Dong2016Compression} by considering four compression quality settings $Q$ = 10, 20, 30 and 40 for the JPEG encoder.
Both JPEG encoder and JPEG image artifacts removal are only applied on the Y channel~\cite{Dong2016Compression}.

\subsubsection{Network training}

A MWCNN model is learned for each degradation setting.
The network parameters are initialized based on the method described in \cite{he2016deep}.
We use the ADAM algorithm \cite{kingma2014adam} with $\alpha = 0.01$, $\beta_1=0.9$, $\beta_2=0.999$ and $\epsilon=10^{-8}$ for optimizing and a mini-batch size of 24.
As to the other hyper-parameters of ADAM, the default setting is adopted.
The learning rate is decayed exponentially from 0.001 to 0.0001 in the 40 epochs.
Rotation or/and flip based data augmentation is used during mini-batch learning.
We use the MatConvNet package \cite{vedaldi2015matconvnet} with cuDNN 6.0 to train our MWCNN.
All the experiments are conducted in the Matlab (R2016b) environment running on a PC with Intel(R) Core(TM) i7-5820K CPU 3.30GHz and an Nvidia GTX1080 GPU.
The learning algorithm converges very fast and it takes about two days to train a MWCNN model.

\begin{table*}[!htbp]\footnotesize

\centering
\vspace{-0.1in}
\caption{Average PSNR(dB)/SSIM results of the competing methods for image denoising with noise levels $\sigma = $ 15, 25 and 50 on datasets Set14, BSD68 and Urban100. Red color indicates the best performance.}
\scalebox{0.885}{
\begin{tabular}{|p{1.25cm}<{\centering}|p{1.28cm}<{\centering}|p{1.88cm}<{\centering}|p{1.88cm}<{\centering}|p{1.88cm}<{\centering}|p{1.88cm}<{\centering}|p{1.88cm}<{\centering}|p{1.88cm}<{\centering}|p{1.88cm}<{\centering}|p{1.88cm}<{\centering}|}
 \hline

Dataset &   $\sigma$ & BM3D~\cite{dabov2007image} & TNRD~\cite{Chen2015Trainable} & DnCNN~\cite{Zhang2016Beyond} & IRCNN~\cite{zhang2017learning} & RED30~\cite{Mao2016Image}  & MemNet~\cite{tai2017memnet} &      MWCNN       \\ \hline
 \hline
 \multirow{3}{*}{Set12}
&15 & 32.37 / 0.8952 & 32.50 / 0.8962 & 32.86 / 0.9027 & 32.77 / 0.9008 &  -  &  - & \textcolor[rgb]{1,0,0}{33.15} / \textcolor[rgb]{1,0,0}{0.9088} \\
&25 & 29.97 / 0.8505 & 30.05 / 0.8515 & 30.44 / 0.8618 & 30.38 / 0.8601 &  -  &  -  & \textcolor[rgb]{1,0,0}{30.79} / \textcolor[rgb]{1,0,0}{0.8711} \\
&50 & 26.72 / 0.7676 & 26.82 / 0.7677 & 27.18 / 0.7827 & 27.14 / 0.7804 & 27.34 / 0.7897 & 27.38 / 0.7931  & \textcolor[rgb]{1,0,0}{27.74} / \textcolor[rgb]{1,0,0}{0.8056} \\ \hline
 \multirow{3}{*}{BSD68}
&15 & 31.08 / 0.8722 & 31.42 / 0.8822 & 31.73 / 0.8906 & 31.63 / 0.8881 &  -  & -   & \textcolor[rgb]{1,0,0}{31.86} / \textcolor[rgb]{1,0,0}{0.8947} \\
&25 & 28.57 / 0.8017 & 28.92 / 0.8148 & 29.23 / 0.8278 & 29.15 / 0.8249 &  -  & -   & \textcolor[rgb]{1,0,0}{29.41} / \textcolor[rgb]{1,0,0}{0.8360} \\
&50 & 25.62 / 0.6869 & 25.97 / 0.7021 & 26.23 / 0.7189 & 26.19 / 0.7171 &  26.35 / 0.7245  &  26.35 / 0.7294 & \textcolor[rgb]{1,0,0}{26.53} / \textcolor[rgb]{1,0,0}{0.7366} \\ \hline
 \multirow{3}{*}{Urban100}
&15 & 32.34 / 0.9220 & 31.98 / 0.9187 & 32.67 / 0.9250 & 32.49 / 0.9244 & -  & -  & \textcolor[rgb]{1,0,0}{33.17} / \textcolor[rgb]{1,0,0}{0.9357} \\
&25 & 29.70 / 0.8777 & 29.29 / 0.8731 & 29.97 / 0.8792 & 29.82 / 0.8839 & -  & -  & \textcolor[rgb]{1,0,0}{30.66} / \textcolor[rgb]{1,0,0}{0.9026} \\
&50 & 25.94 / 0.7791 & 25.71 / 0.7756 & 26.28 / 0.7869 & 26.14 / 0.7927 &  26.48 / 0.7991  & 26.64 / 0.8024  & \textcolor[rgb]{1,0,0}{27.42} / \textcolor[rgb]{1,0,0}{0.8371} \\ \hline
\end{tabular}
}
\label{tab:denoising}
\vspace{-0.07in}
\end{table*}

\begin{table*}[!htbp]\footnotesize
\centering
\caption{Average PSNR(dB) / SSIM results of  the competing methods for SISR with scale factors $S=$ 2, 3 and 4 on datasets Set5, Set14, BSD100 and Urban100. Red color indicates the best performance.}
\scalebox{0.74}{
\begin{tabular}{|p{1.2cm}<{\centering}|p{0.7cm}<{\centering}|p{1.62cm}<{\centering}|p{1.62cm}<{\centering}|p{1.62cm}<{\centering}|p{1.62cm}<{\centering}|p{1.62cm}<{\centering}|p{1.62cm}<{\centering}|p{1.62cm}<{\centering}|p{1.62cm}<{\centering}|p{1.77cm}<{\centering}|p{1.62cm}<{\centering}|p{1.62cm}<{\centering}|} \hline

Dataset& $S$ & RCN~\cite{Shi2017Structure} & VDSR~\cite{kim2015accurate}  & DnCNN~\cite{Zhang2016Beyond} & RED30~\cite{Mao2016Image} & SRResNet~\cite{Ledig2017Photo} & LapSRN~\cite{lai2017deep} & DRRN~\cite{Tai2017Image}   & MemNet~\cite{tai2017memnet} & WaveResNet~\cite{bae2017beyond} & MWCNN \\  \hline
\hline
 \multirow{3}{*}{Set5}
 &$\times$2    & 37.17 / 0.9583   & 37.53 / 0.9587     &   37.58 / 0.9593  & 37.66 / 0.9599 &       -        & 37.52 / 0.9590 & 37.74 / 0.9591 & 37.78 / 0.9597   & 37.57 / 0.9586 & \textcolor[rgb]{1,0,0}{37.91} / \textcolor[rgb]{1,0,0}{0.9600} \\
 &$\times$3    & 33.45 / 0.9175   &   33.66 / 0.9213   &   33.75 / 0.9222  & 33.82 / 0.9230 &       -        &      -         & 34.03 / 0.9244 & 34.09 / 0.9248  & 33.86 / 0.9228  & \textcolor[rgb]{1,0,0}{34.17} / \textcolor[rgb]{1,0,0}{0.9271} \\
 &$\times$4    & 31.11 / 0.8736   &   31.35 / 0.8838   &   31.40 / 0.8845  & 31.51 / 0.8869 & 32.05 / 0.8902 & 31.54 / 0.8850 & 31.68 / 0.8888 &  31.74 / 0.8893 & 31.52 / 0.8864   & \textcolor[rgb]{1,0,0}{32.12} / \textcolor[rgb]{1,0,0}{0.8941} \\
  \hline
   \multirow{3}{*}{Set14}
 &$\times$2    & 32.77 / 0.9109   & 33.03 / 0.9124   &   33.04 / 0.9118  & 32.94 / 0.9144 &        -        & 33.08 / 0.9130 & 33.23 / 0.9136 &  33.28 / 0.9142   & 33.09 / 0.9129  & \textcolor[rgb]{1,0,0}{33.70} / \textcolor[rgb]{1,0,0}{0.9182} \\
 &$\times$3    & 29.63 / 0.8269   & 29.77 / 0.8314   &   29.76 / 0.8349  & 29.61 / 0.8341 &        -        &       -        & 29.96 / 0.8349 & 30.00 / 0.8350  &  29.88 / 0.8331  & \textcolor[rgb]{1,0,0}{30.16} /  \textcolor[rgb]{1,0,0}{0.8414}  \\
 &$\times$4    & 27.79 / 0.7594  & 28.01 / 0.7674   &   28.02 / 0.7670  & 27.86 / 0.7718 &  \textcolor[rgb]{1,0,0}{28.49} / 0.7783 & 28.19 / 0.7720 & 28.21 / 0.7720 &  28.26 / 0.7723  & 28.11 / 0.7699  &  28.41 /  \textcolor[rgb]{1,0,0}{0.7816} \\ \hline
 \multirow{3}{*}{BSD100}
 &$\times$2   &  - & 31.90 / 0.8960   &   31.85 / 0.8942  & 31.98 / 0.8974 & - &  31.80 / 0.8950  & 32.05 / 0.8973 &  32.08 / 0.8978 & 32.15 / 0.8995 & \textcolor[rgb]{1,0,0}{32.23} / \textcolor[rgb]{1,0,0}{0.8999} \\
 &$\times$3   &  - & 28.82 / 0.7976   &   28.80 / 0.7963  & 28.92 / 0.7993 & - & - & 28.95 / 0.8004 & 28.96 / 0.8001& 28.86 / 0.7987  &\textcolor[rgb]{1,0,0}{29.12} / \textcolor[rgb]{1,0,0}{0.8060} \\
 &$\times$4   &  - & 27.29 / 0.7251   &   27.23 / 0.7233  & 27.39 / 0.7286 & 27.56 / 0.7354 & 27.32 / 0.7280 & 27.38 / 0.7284 & 27.40 / 0.7281 & 27.32 / 0.7266 & \textcolor[rgb]{1,0,0}{27.62} / \textcolor[rgb]{1,0,0}{0.7355} \\
\hline
\multirow{3}{*}{Urban100}
 &$\times$2   & - &30.76 / 0.9140 &     30.75 / 0.9133 & 30.91 / 0.9159 & - & 30.41 / 0.9100 & 31.23 / 0.9188 & 31.31 / 0.9195 & 30.96 / 0.9169 &\textcolor[rgb]{1,0,0}{32.30} / \textcolor[rgb]{1,0,0}{0.9296} \\
 &$\times$3   & - &27.14 / 0.8279 &     27.15 / 0.8276 & 27.31 / 0.8303 & - & - &  27.53 / 0.8378 & 27.56 / 0.8376  & 27.28 / 0.8334  &\textcolor[rgb]{1,0,0}{28.13} / \textcolor[rgb]{1,0,0}{0.8514} \\
 &$\times$4   & - &25.18 / 0.7524 &     25.20 / 0.7521 & 25.35 / 0.7587 & 26.07 / 0.7839 & 25.21 / 0.7560 & 25.44 / 0.7638 & 25.50 / 0.7630   & 25.36 / 0.7614  & \textcolor[rgb]{1,0,0}{26.27} / \textcolor[rgb]{1,0,0}{0.7890} \\
\hline


\end{tabular}
}
\vspace{-0.07in}
\label{tab:sr}
\end{table*}

\begin{table*}[!htbp]\footnotesize
\centering
\caption{Average PSNR(dB) / SSIM results of the competing methods for JPEG image artifacts removal with quality factors $Q = $ 10, 20, 30 and 40 on datasets Classic5 and LIVE1. Red color indicates the best performance.}
\scalebox{0.866}{
\begin{tabular}{|p{1.63cm}<{\centering}|p{1.63cm}<{\centering}|p{2.22cm}<{\centering}|p{2.22cm}<{\centering}|p{2.22cm}<{\centering}|p{2.22cm}<{\centering}|p{2.22cm}<{\centering}|p{2.22cm}<{\centering}|p{2.22cm}<{\centering}|p{1.62cm}<{\centering}|}
 \hline

Dataset & $Q$ & JPEG & ARCNN~\cite{Dong2016Compression} & TNRD~\cite{Chen2015Trainable}   & DnCNN~\cite{Zhang2016Beyond}    & MemNet~\cite{tai2017memnet}   &   MWCNN       \\ \hline
 \hline
 \multirow{4}{*}{Classic5}
&10 & 27.82 / 0.7595 & 29.03 / 0.7929 & 29.28 / 0.7992 & 29.40 / 0.8026 &   29.69 / 0.8107 & \textcolor[rgb]{1,0,0}{30.01} / \textcolor[rgb]{1,0,0}{0.8195} \\
&20 & 30.12 / 0.8344 & 31.15 / 0.8517 & 31.47 / 0.8576 & 31.63 / 0.8610 &   31.90 / 0.8658 & \textcolor[rgb]{1,0,0}{32.16} / \textcolor[rgb]{1,0,0}{0.8701}  \\
&30 & 31.48 / 0.8744 & 32.51 / 0.8806 & 32.78 / 0.8837 & 32.91 / 0.8861 &  - & \textcolor[rgb]{1,0,0}{33.43} / \textcolor[rgb]{1,0,0}{0.8930}  \\
&40 & 32.43 / 0.8911 & 33.34 / 0.8953 &       -        & 33.77 / 0.9003 &  - & \textcolor[rgb]{1,0,0}{34.27} / \textcolor[rgb]{1,0,0}{0.9061}  \\ \hline
\multirow{4}{*}{LIVE1}
&10 & 27.77 / 0.7730 & 28.96 / 0.8076 & 29.15 / 0.8111 & 29.19 / 0.8123 &  29.45 / 0.8193 & \textcolor[rgb]{1,0,0}{29.69} / \textcolor[rgb]{1,0,0}{0.8254} \\
&20 & 30.07 / 0.8512 & 31.29 / 0.8733 & 31.46 / 0.8769 & 31.59 / 0.8802 &  31.83 / 0.8846 & \textcolor[rgb]{1,0,0}{32.04} / \textcolor[rgb]{1,0,0}{0.8885}  \\
&30 & 31.41 / 0.9000 & 32.67 / 0.9043 & 32.84 / 0.9059 & 32.98 / 0.9090 & -  & \textcolor[rgb]{1,0,0}{33.45} / \textcolor[rgb]{1,0,0}{0.9153}\\
&40 & 32.35 / 0.9173 & 33.63 / 0.9198 &       -        & 33.96 / 0.9247 & -  & \textcolor[rgb]{1,0,0}{34.45} / \textcolor[rgb]{1,0,0}{0.9301}  \\ \hline
\end{tabular}
}
\label{tab:deblock}
\vspace{-0.12in}
\end{table*}

\subsection{Quantitative and qualitative evaluation}

{\color{black}
%
%
%
In this subsection, all the MWCNN models use the same network setting described in Sec.~\ref{sec3.2}, and 2D Haar wavelet is adopted.}
%
%


\subsubsection{Image denoising}

\begin{figure*}[!htbp]
\setlength{\abovecaptionskip}{0pt}
\setlength{\belowcaptionskip}{0pt}
\begin{center}
\vspace{-1.9ex}
  \includegraphics[width=1\textwidth]{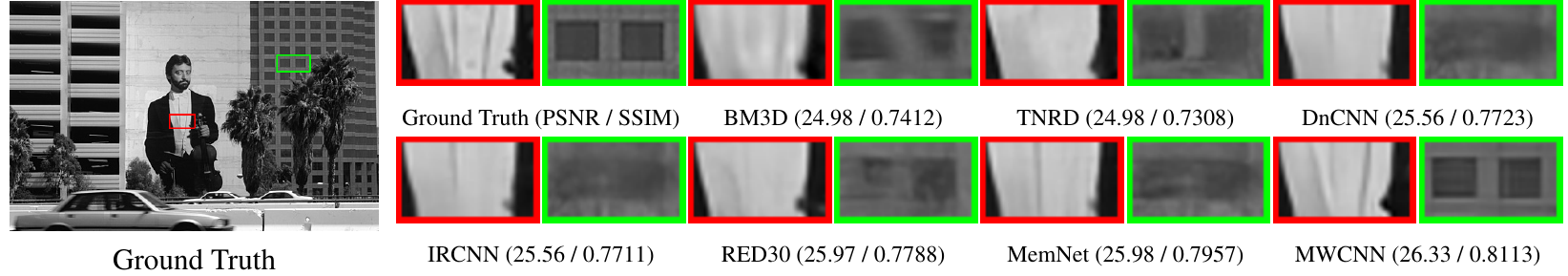}\\
  \vspace{-1ex}
\end{center}
\vspace{-2.2ex}
\caption{\emph{Image denoising} results of ``$Test011$'' (BSD68) with noise level 50. }\label{fig:DN50_test044}
\vspace{-0.13in}
\end{figure*}

Except CBM3D~\cite{dabov2007image} and CDnCNN~\cite{Zhang2016Beyond}, most denoising methods are only tested on gray images.
Thus, we train our MWCNN by using the gray images, and compare with six competing denoising methods, \ie, BM3D~\cite{dabov2007image}, TNRD~\cite{Chen2015Trainable},  DnCNN~\cite{Zhang2016Beyond}, IRCNN~\cite{zhang2017learning}, RED30~\cite{Mao2016Image}, and MemNet~\cite{tai2017memnet}.
We evaluate the denoising methods on three test datasets, \ie, Set12~\cite{Zhang2016Beyond}, BSD68~\cite{MartinFTM01}, and Urban100~\cite{huang2015single}.
Table \ref{tab:denoising} lists the average PSNR/SSIM results of the competing methods on these three datasets.
We note that our MWCNN only slightly outperforms DnCNN by about $0.1\sim0.3$dB in terms of PSNR on BSD68.
%
%
As to other datasets, our MWCNN generally achieves favorable performance when compared with the competing methods.
When the noise level is high (\eg, $\sigma = 50$), the average PSNR by our MWCNN can be 0.5dB higher than that by DnCNN on Set12, and 1.2dB higher on Urban100.
Figure~\ref{fig:DN50_test044} shows the denoising results of the images \emph{Test011} from Set68 with the noise level $\sigma = 50$.
One can see that our MWCNN is promising in recovering image details and structures, and can obtain visually more pleasant result than the competing methods.
Please refer to the supplementary materials for more results on Set12 and Urban100.


\subsubsection{Single image super-resolution}

Following \cite{kim2015accurate}, SISR is only applied to the luminance channel, \ie Y in YCbCr color space.
We test MWCNN on four datasets, \ie, Set5~\cite{bevilacqua2012low}, Set14~\cite{zeyde2010single}, BSD100~\cite{MartinFTM01}, and Urban100~\cite{huang2015single}, because they are widely adopted to evaluate SISR performance.
Our MWCNN is compared with eight CNN-based SISR methods, including RCN~\cite{Shi2017Structure}, VDSR~\cite{kim2015accurate}, DnCNN~\cite{Zhang2016Beyond}, RED30~\cite{Mao2016Image}, SRResNet~\cite{Ledig2017Photo}, LapSRN~\cite{lai2017deep}, DRRN~\cite{Tai2017Image}, and MemNet~\cite{tai2017memnet}.
Due to the source code of SRResNet is not released, its results are from~\cite{Ledig2017Photo} and are incomplete.

\begin{figure*}[!htbp]
\setlength{\abovecaptionskip}{0pt}
\setlength{\belowcaptionskip}{0pt}
 \begin{center}
\vspace{-0.5ex}
  \includegraphics[width=1\textwidth]{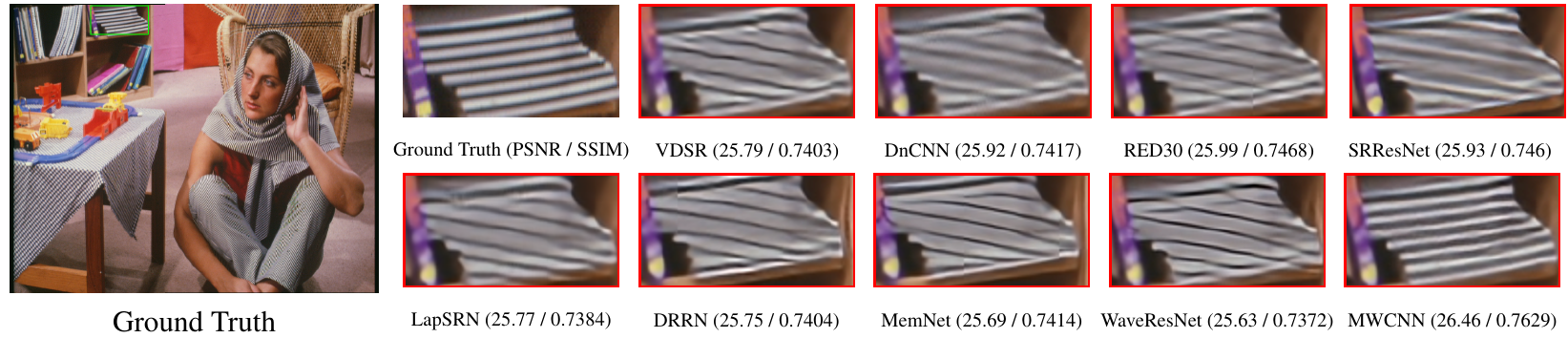}\\
  \vspace{-1ex}
\end{center}
\vspace{-2.1ex}
\caption{\emph{Single image super-resolution} results of ``$barbara$'' (Set14) with upscaling factor $\times$4. }\label{fig:srs4_barbara}
\vspace{-0.13in}
\end{figure*}

Table \ref{tab:sr} lists the average PSNR/SSIM results of the competing methods on the four datasets.
Our MWCNN performs favorably in terms of both PSNR and SSIM indexes.
Compared with VDSR, our MWCNN achieves a notable gain of about 0.4dB by PSNR on Set5 and Set14.
On Urban100, our MWCNN outperforms VDSR by about 0.9$\sim$1.4dB.
Obviously, WaveResNet \etal~\cite{bae2017beyond} sightly outperform VDSR, and also is still inferior to MWCNN.
We note that the network depth of SRResNet is 34, while that of MWCNN is 24.
Moreover, SRResNet is trained with a much larger training set than MWCNN.
Even so, when the scale factor is 4, MWCNN achieve slightly higher PSNR values on Set5 and BSD100, and is comparable to SRResNet on Set14.
Figure~\ref{fig:srs4_barbara} shows the visual comparisons of the competing methods on the images \emph{Barbara} from Set14.
Thanks to the frequency and location characteristics of DWT, our MWCNN can correctly recover the fine and detailed textures, and produce sharp edges.
{\color{black} Furthermore, for Track 1 of NTIRE 2018 SR challenge ($\times8$ SR)~\cite{NTIRE2018}, our improved MWCNN is lower than the Top-1 method by 0.37dB}.

\begin{figure*}[!htbp]
\setlength{\abovecaptionskip}{0pt}
\setlength{\belowcaptionskip}{0pt}
 \begin{center}
\vspace{-0.5ex}
  \includegraphics[width=1\textwidth]{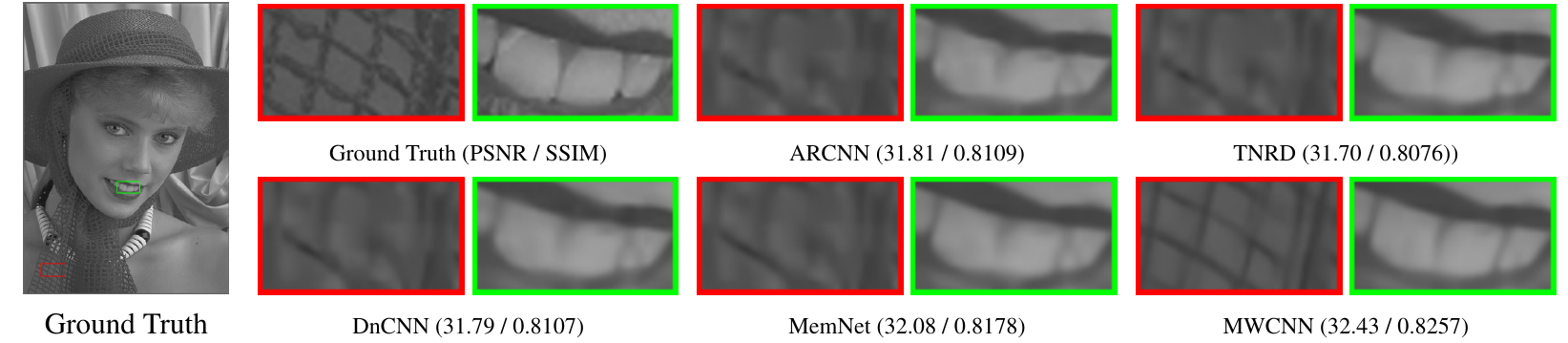}\\
  \vspace{-1ex}
\end{center}
 \vspace{-2.1ex}
\caption{\emph{JPEG image artifacts removal} results of ``$womanhat$'' (LIVE1) with quality factor 10. }\label{fig:db10_building}
\vspace{-0.15in}
\end{figure*}

\subsubsection{JPEG image artifacts removal}

In JPEG compression, an image is divided into non-overlapped $8\times 8$ blocks.
Discrete cosine transform (DCT) and quantization are then applied to each block, thus introducing the blocking artifact.
The quantization is determined by a quality factor $Q$ to control the compression rate.
Following \cite{Dong2016Compression}, we consider four settings on quality factor, \eg, $Q$ = 10, 20, 30 and 40, for the JPEG encoder.
Both JPEG encoder and JPEG image artifacts removal are only applied to the Y channel.
In our experiments, MWCNN is compared with four competing methods, \ie, ARCNN~\cite{Dong2016Compression}, TNRD~\cite{Chen2015Trainable}, DnCNN~\cite{Zhang2016Beyond}, and MemNet~\cite{tai2017memnet} on the two datasets, {\color{black}\ie, Classic5 and LIVE1~\cite{moorthy2009visual}}.
We do not consider~\cite{Guo2016Building,Guo2017One} due to their source codes are unavailable.

Table~\ref{tab:deblock} lists the average PSNR/SSIM results of the competing methods on Classic5 and LIVE1.
%
%
For any of the four quality factors, our MWCNN performs favorably in terms of quantitative metrics on the two datasets.
On Classic5 and LIVE1, the PSNR values of MWCNN can be 0.2$\sim$0.3dB higher than those of the second best method (\ie, MemNet~\cite{tai2017memnet}) for the quality factor of 10 and 20.
Figure~\ref{fig:db10_building} shows the results on the image \emph{womanhat} from LIVE1 with the quality factor 10.
One can see that MWCNN is effective in restoring detailed textures and sharp salient edges.



%
%
%
%

\subsubsection{Run time}

%
Table~\ref{tab:runtime} lists the GPU run time of the competing methods for the three tasks.
The Nvidia cuDNN-v6.0 deep learning library is adopted to accelerate the GPU computation under Ubuntu 16.04 system.
Specifically, only the CNN-based methods with source codes are considered in the comparison.
For three tasks, the run time of MWCNN is far less than several state-of-the-art methods, including RED30~\cite{Mao2016Image}, MemNet~\cite{Tai2017Image} and DRRN~\cite{Tai2017Image}.
Note that the three methods also perform poorer than MWCNN in terms of PSNR/SSIM metrics.
In comparison to the other methods, MWCNN is moderately slower by speed but can achieve higher PSNR/SSIM indexes.
%
%
%
The result indicates that, instead of the increase of network depth/width, the effectiveness of MWCNN should be attributed to the incorporation of CNN and DWT.

\begin{table}[!htbp]\footnotesize
\centering
\caption{Run time (in seconds) of the competing methods for the three tasks on images of size 256$\times$256, 512$\times$512 and 1024$\times$1024:
image denosing is tested on noise level 50, SISR is tested on scale $\times$2, and JPEG image deblocking is tested on quality factor 10.}
\scalebox{0.79}{
\begin{tabular}{|p{1.29cm}<{\centering}|p{1.4cm}<{\centering}|p{1.35cm}<{\centering}|p{1.35cm}<{\centering}|p{1.4cm}<{\centering}|p{1.05cm}<{\centering}|}
\hline
  \multicolumn{6}{|c|}{\textbf{Image Denoising}} \\ \hline
 \hline
Size & TNRD~\cite{Chen2015Trainable} & DnCNN~\cite{Zhang2016Beyond} & RED30~\cite{Mao2016Image} &  MemNet~\cite{Tai2017Image}   &  MWCNN          \\ \hline
  256$\times$256    & 0.010  &   0.0143   &   1.362    &   0.8775  &   0.0586  \\ \hline
  512$\times512$    & 0.032  &   0.0487   &   4.702    &   3.606   &   0.0907  \\ \hline
  1024$\times 1024$ & 0.116  &   0.1688   &   15.77    &   14.69   &   0.3575   \\ \hline \hline
  \multicolumn{6}{|c|}{\textbf{Single Image Super-Resolution}} \\ \hline
  \hline
Size &  VDSR~\cite{kim2015accurate} & LapSRN~\cite{lai2017deep} & DRRN~\cite{Tai2017Image} & MemNet~\cite{Mao2016Image}  &  MWCNN          \\ \hline
  256$\times$256    & 0.0172 &  0.0229    &    3.063   &   0.8774   &   0.0424  \\ \hline
  512$\times512$    & 0.0575 &  0.0357    &    8.050   &   3.605   &   0.0780    \\ \hline
  1024$\times 1024$ & 0.2126 &  0.1411    &    25.23   &   14.69    &   0.3167    \\ \hline
    \hline
   \multicolumn{6}{|c|}{\textbf{JPEG Image Artifacts Removal}} \\ \hline
   \hline
Size &   ARCNN~\cite{Dong2016Compression} & TNRD~\cite{Chen2015Trainable} & DnCNN~\cite{Zhang2016Beyond} & MemNet~\cite{Mao2016Image}  &  MWCNN          \\ \hline
  256$\times$256     &  0.0277  & 0.009   &   0.0157  &   0.8775   &   0.0531  \\ \hline
  512$\times512$     &  0.0532  & 0.028   &   0.0568  &   3.607    &   0.0811    \\ \hline
  1024$\times 1024$  &  0.1613  & 0.095   &   0.2012  &   14.69    &   0.2931    \\ \hline

\end{tabular}
}
\label{tab:runtime}
\vspace{-0.145in}
\end{table}


\begin{table*}[!htbp]\footnotesize
\centering
\vspace{-0.00in}
\caption{
{\color{black} Performance comparison in terms of average PSNR (dB) and run time (in seconds): image denosing is tested on noise level 50 and JPEG image deblocking is tested on quality factor 10.}
}
\scalebox{0.73}{
\begin{tabular}{|p{0.99cm}<{\centering}|p{1.70cm}<{\centering}|p{1.70cm}<{\centering}|p{1.70cm}<{\centering}|p{1.70cm}<{\centering}|p{1.8cm}<{\centering}|p{1.8cm}<{\centering}|p{1.92cm}<{\centering}|p{1.92cm}<{\centering}|p{1.90cm}<{\centering}|p{1.90cm}<{\centering}|}
 \hline
 Dataset &      Dilated~\cite{yu2015multi} &      Dilated-2  & U-Net~\cite{Ronneberger2015U} & U-Net+S  & U-Net+D & DCF~\cite{Han2017Framing} & WaveResNet~\cite{bae2017beyond} &  MWCNN (Haar) &   MWCNN (DB2)  &   MWCNN (HD)      \\ \hline  \hline

   \multicolumn{11}{|c|}{\textbf{Image Denoising ($\sigma=50$)}} \\
 \hline

Set12    & 27.45 / 0.181 & 24.81 / 0.185 & 27.42 / 0.079 &  27.41 / \textbf{0.074}  & 27.46 / 0.080 &  27.38 /  0.081 &  27.49 / 0.179  & 27.74 / 0.078   & \textbf{27.77} / 0.134 &  27.73 / 0.101 \\ \hline
BSD68    & 26.35 / 0.142 & 24.32 / 0.174 & 26.30 / 0.076 &  26.29 / \textbf{0.071} & 26.21 / 0.075 &  26.30 /  0.075 &  26.38 / 0.143  & 26.53 / 0.072   & \textbf{26.54} / 0.122 &  26.52 / 0.088  \\ \hline
Urban100 & 26.56 / 0.764 & 24.18 / 0.960 & 26.68 / 0.357 &  26.72 / \textbf{0.341}  & 26.99 / 0.355 &  26.65 / 0.354  &  - / -  & 27.42 / 0.343   & \textbf{27.48} / 0.634 &  27.35 / 0.447  \\ \hline
 \hline
   \multicolumn{11}{|c|}{\textbf{JPEG Image Artifacts Removal (PC=10)}} \\ \hline
 \hline
Classic5   & 29.72 / 0.287 &  29.49 / 0.302 & 29.61 / 0.093  & 29.60 / \textbf{0.082} & 29.68 / 0.097 & 29.57 / 0.104 &   - / -  &  30.01 / 0.088 & \textbf{30.04} / 0.195 &  29.97 / 0.136 \\ \hline
LIVE1      & 29.49 / 0.354 &  29.26 / 0.376 & 29.36 / 0.112  & 29.36 / \textbf{0.109} & 29.43 / 0.120 & 29.38 / 0.155 &   - / -  &  29.69 / 0.112  & \textbf{29.70} / 0.265 &  29.66 / 0.187 \\ \hline

\end{tabular}
}
\label{tab:denoising_part}
\vspace{-0.13in}
\end{table*}

\subsection{Comparison of MWCNN variants}

{\color{black}
Using image denoising and JPEG image artifacts as examples,} we compare the PSNR results by three MWCNN variants, including: (i) MWCNN (Haar): the default MWCNN with Haar wavelet, (ii) MWCNN (DB2): MWCNN with \emph{Daubechies-2} wavelet, and (iii) MWCNN (HD): MWCNN with Haar in contracting subnetwork and \emph{Daubechies-2} in expanding subnetwork.
{\color{black} Then, ablation experiments are provided for verifying the effectiveness of additionally embedded wavelet: (i) the default U-Net with same architecture to MWCNN, (ii) U-Net+S: using sum connection instead of concatenation, and (iii) U-Net+D: adopting learnable conventional downsamping filters instead of Max pooling.
Two 24-layer dilated CNNs are also considered: (i) Dilated: the hybrid dilated convolution~\cite{Wang2017Understanding} to suppress the gridding effect, and (ii) Dilated-2: the dilate factor  of all layers is set to 2.
}
%
The WaveResNet method in~\cite{bae2017beyond} is provided to be compared.
Moreover, due to its code is unavailable, a self-implementation of deep convolutional framelets (DCF)~\cite{Ye2017Deep} is also considered in the experiments.
%

%
Table~\ref{tab:runtime} lists the PSNR and run time results of these methods.
And we have the following observations.
{\color{black}
(i) The gridding effect with the sparse sampling and inconsistence of local information authentically has adverse influence on restoration performance.
(ii) The ablation experiments indicate that using sum connection instead of concatenation can improve efficiency without decreasing PNSR.
%
 Due to the special group of filters with the biorthogonal and time-frequency localization property in wavelet, our embedded wavelet own more puissant ability for image restoration than pooling operation and learnable downsamping filters.
The worse performance of DCF also indicates that independent processing of subbands harms final result.
(iii) Compared to MWCNN (DB2) and MWCNN (HD), using Haar wavelet for downsampling and upsampling in network is the best choice in terms of quantitative and qualitative evaluation.}
MWCNN (Haar) has similar run time with dilated CNN and U-Net but achieves higher PSNR results, which demonstrates the effectiveness of MWCNN for tradeoff between performance and efficiency.
%
%
%
%
%
%
%
%
%

Note that our MWCNN is quite different with DCF~\cite{Ye2017Deep}:
DCF incorporates CNN with DWT in the view of decomposition, where different CNNs are deployed to each subband.
However, the results in Table~\ref{tab:denoising_part} indicates that independent processing of subbands is not suitable for image restoration.
%
On the contrary, MWCNN combines DWT to CNN from perspective of enlarging receptive field without information loss, allowing to embed DWT with any CNNs with pooling.
{\color{black}Moreover, our embedded DWT can be treated as predefined parameters to ease network learning, and the dynamic range of subbands can be jointly adjusted by the CNN blocks.}
Taking all subbands as input, MWCNN is more powerful in modeling inter-band dependency.
%

%
%
%



\begin{table}[!htbp]\footnotesize
\centering
\caption{Average PSNR (dB) and run time (in seconds) of MWCNNs with different levels on Gaussian denoising with the noise level of 50.}
\scalebox{0.83}{
\begin{tabular}{|p{1.8cm}<{\centering}|p{1.5cm}<{\centering}|p{1.5cm}<{\centering}|p{1.5cm}<{\centering}|p{1.5cm}<{\centering}|p{1.05cm}<{\centering}|}
 \hline

Dataset & MWCNN-1 & MWCNN-2 & MWCNN-3 & MWCNN-4           \\ \hline \hline
Set12      &   27.14 / 0.047   &    27.62 / 0.068    &    27.74 / 0.082     &    27.74 / 0.091  \\ \hline
BSD68      &   26.16 / 0.044   &    26.45 / 0.063    &    26.53 / 0.074     &    26.54 / 0.084  \\ \hline
Urban100   &   26.08 / 0.212   &    27.10 / 0.303    &    27.42 / 0.338     &    27.44 / 0.348  \\ \hline
\end{tabular}
}
\label{tab:depth}
\vspace{-0.13in}
\end{table}

As described in Sec.~\ref{sec3.2}, our MWCNN can be extended to higher level of wavelet decomposition.
Nevertheless, higher level inevitably results in deeper network and heavier computational burden.
Thus, a suitable level is required to balance efficiency and performance.
Table~\ref{tab:depth} reports the PSNR and run time results of MWCNNs with the levels of 1 to 4 (\ie, MWCNN-1 $\sim$ MWCNN-4).
It can be observed that MWCNN-3 {\color{black} with 24-layer architecture} performs much better than MWCNN-1 and MWCNN-2, while MWCNN-4 only performs negligibly better than MWCNN-3 in terms of the PSNR metric.
Moreover, the speed of MWCNN-3 is also moderate compared with other levels.
%
Taking both efficiency and performance gain into account, we choose MWCNN-3 as the default setting.


%
%
%
%



\section{Conclusion}\label{sec:con}

This paper presents a multi-level wavelet-CNN (MWCNN) architecture for image restoration, which consists of a contracting subnetwork and a expanding subnetwork.
The contracting subnetwork is composed of multiple levels of DWT and CNN blocks, while the expanding subnetwork is composed of multiple levels of IWT and CNN blocks.
Due to the invertibility, frequency and location property of DWT, MWCNN is safe to perform subsampling without information loss, and is effective in recovering detailed textures and sharp structures from degraded observation.
As a result, MWCNN can enlarge receptive field with better tradeoff between efficiency and performance.
Extensive experiments demonstrate the effectiveness and efficiency of MWCNN on three restoration tasks, \ie, image denoising, SISR, and JPEG compression artifact removal.

In future work, we will extend MWCNN for more general restoration tasks such as image deblurring and blind deconvolution.
Moreover, our MWCNN can also be used to substitute the pooling operation in the CNN architectures for high-level vision tasks such as image classification.

\section*{Acknowledgement}
\noindent
This work is partially supported by the National Science Foundation of China (NSFC) (grant No.s 61671182 and 61471146).
{\small
\bibliographystyle{ieee}
\bibliography{egbib}
}

\end{document}